\documentclass{elsarticle}

\usepackage{changepage}

\usepackage[utf8]{inputenc} 
\usepackage[T1]{fontenc}    

\usepackage[margin=3.3cm]{geometry}

\usepackage[parfill]{parskip} 

\usepackage{graphicx}
\usepackage{url}            
\usepackage{booktabs}       
\usepackage{amsfonts}       
\usepackage{nicefrac}       
\usepackage{microtype}      

\usepackage[super]{nth}
\usepackage{amsmath}

\usepackage{multirow}
\usepackage{rotating}
\usepackage{algorithm}
\usepackage{algpseudocode}
\usepackage{tikz}
\usetikzlibrary{fit, positioning, shapes,snakes}
\usepackage{makecell}

\usepackage{etoolbox}
\apptocmd{\thebibliography}{\setlength{\itemsep}{18pt}}{}{}

\title{Mixture Density Conditional Generative Adversarial Network Models (MD-CGAN)}

\author{
 Jaleh Zand \& Stephen Roberts\\
 Machine Learning Research Group \& 
 Oxford-Man Institute of Quantitative Finance\\ University of Oxford\\
 \{jz,sjrob\}@robots.ox.ac.uk
}

\makeatletter
\def\ps@pprintTitle{%
   \let\@oddhead\@empty
   \let\@evenhead\@empty
   \def\@oddfoot{\reset@font\hfil\thepage\hfil}
   \let\@evenfoot\@oddfoot
}
\makeatother

\begin{document}

\begin{frontmatter}

\begin{abstract}
Generative Adversarial Networks (GANs) have gained significant attention in recent years, with impressive applications highlighted in computer vision in particular.  Compared to such examples, however, there have been more limited applications of GANs to time series modelling, including forecasting.  In this work, we present the \emph{Mixture Density Conditional Generative Adversarial Model} (MD-CGAN), with a focus on time series forecasting.  We show that our model is capable of estimating a probabilistic posterior distribution over forecasts and that, in comparison to a set of benchmark methods, the MD-CGAN model performs well, particularly in situations where noise is a significant component of the observed time series.  Further, by using a Gaussian mixture model as the output distribution,  MD-CGAN offers posterior predictions that are non-Gaussian.
\end{abstract}

\end{frontmatter}

\section{Introduction}\label{intro}

\emph{Generative Adversarial Networks} (GANs) have been one of the many breakthroughs in Deep Learning methods in recent years. Several different variations of the model have been introduced since the method was first introduced in \cite{GAN}. One of the most popular variations of this work is that of the \emph{Conditional Generative Adversarial Network} (CGAN) \cite{CGAN} in which the \emph{generator} and \emph{discriminator} (we briefly review the GAN process in Section \ref{ganmodel}) are both conditioned on some observed information.  In the application of time series forecasting, future values are conditioned on information observed from the past, either from the time series itself, some set of associated exogenous data, or a combination of the two. This conditioning addition to the GAN formalism makes the CGAN approach particularly useful as a foundational model for time series prediction.  Most applications of (C)GANs, however, have been within computer vision and, to a lesser extent, in natural language processing and simulation models \cite{s1,s2,s3}.

The literature on the application of any form of GAN model to problems associated with time series is, to date, limited.  However, some recent literature shows the potential usefulness of the method. For example, the work reported in \cite{time-CGAN} applies a recurrent GAN to generate realistic, synthetic, medical data series and in \cite{wiese2020quant} a (standard) GAN model is used to generate realistic financial asset prices and analyse their distributions.

In \cite{highfreq} a GAN is used to forecast high-frequency stock data and in \cite{missing_val} GANs are used to generate missing values in incomplete time series.  We note that the GAN models used in all these applications make \emph{point estimates} for the forecast. Although a perfectly valid approach, and one that has a long history in time series forecasting, we argue that \emph{probabilistic} forecasts are a prerequisite in many application domains, in which knowledge of the predictive uncertainty is as vital as the prediction itself.  In this work, we expand on the CGAN algorithm to allow a full predictive probability distribution, rather than a point value. To obtain richer predictive densities we model the posterior distribution using a finite Gaussian Mixture model (GMM).  Although we find only occasional evidence that such non-Gaussian predictions offer significant benefits, we note that producing them is not much more costly than single Gaussian predictive distributions, and so present our approach as a more general multi-component model. 

\subsection{Related work}
\label{sec:review}
We here briefly review recent literature which is close to our approach.  We start by noting the work of \cite{ganCluster}, in which a mixture of GAN models is proposed as a data clustering model.  Although clearly related, this is somewhat different to our approach, in which we use a single GAN generator, linked to the hyperparameters of the posterior mixture model, rather than a mixture of GAN models. Furthermore, our goal is forecasting rather than unsupervised data classification.  The approaches advocated in \cite{deligan1} and \cite{deligan2} propose a latent space, used for sampling latent vectors in the GAN,  formulated via a Gaussian mixture. The latter replaces, in these papers, the single Gaussian distribution used for such generation in standard GAN models. In both these papers,  the generator and discriminator have a similar structure to a standard GAN.  In \cite{ganDis} a mixture model is used, but for the discriminator alone, with the generator being that of a standard GAN model; we note the difference to our approach, in which the \emph{generator} is a GMM. Finally, \cite{ganNips} compare (standard) GAN models to GMMs for image generation.  The authors show that GANs are superior in their ability to generate sharp images, but note that mixture models offer more efficient inference.  They propose a combination of the two and introduce a GAN model in which the GAN generator is a mixture model.  However, the sample generator still makes point estimates from the multimodal distribution in order to retain a discriminator function the same as that of a standard GAN model.  We offer discriminator extensions which allow the GAN process to operate on the full (multi-modal) posterior distribution.

\subsection{Paper structure}
\label{sec:struct}
The rest of this paper is set out as follows:  in Section \ref{ganmodel} we provide a brief overview of the (C)GAN model, introducing the key concepts.  In Section \ref{model} we present the structure of the MD-CGAN model.  In Section \ref{experiments} we test the model on a variety of `real-world' datasets and discuss the results. Finally, in Section \ref{conclusion}, we conclude.

\section{The GAN and CGAN Model}
\label{ganmodel}
The goal of the GAN model is to estimate a generative model using an \emph{adversarial} process \cite{GAN}.  This is achieved by simultaneously training two models. Firstly, a generative model $G$, that in the case of data forecasting learns past patterns in the data and infers the predictive values. Secondly,  a discriminative model $D$, that determines how likely a sample was to originate from the `true' training data, compared to being sampled from the generator.  The generator is hence matched against an adversary,  the discriminator (whose goal is to detect the difference between a true data sample and one created by the generator).  Components of the model are then trained (via an optimization process) to maximize the probability of the discriminator being unable to distinguish true from generated data samples.  Typically, including the approach we take here, the generator and the discriminator are both constructed as multilayer perceptrons, with stochastic gradient methods being employed to obtain optimization.

We start by formulating the definitions which we use in common across the GAN models we test.  We consider a time series, $y_t$. Our aim is to estimate the forecast of some $y_{t' > t}^{f}$, conditioned on a set of observations which we denote $\mathbf{x}_t$.  The inputs to the generator network are $\mathbf{x}_t$ and $\mathbf{z}_g$, where $\mathbf{z}_g$ is a collection of samples from a normal distribution, $p(z_n)_g = \mathcal{N}(0,\mathrm{var}_{\mathrm{data}})$.  During model training, the output from the generator, $y_{t'}^f$, as well as the true forecast sample $y_{t'}$, are fed to the discriminator, whose role is to discriminate between them i.e. to identify $y_{t'}^f$ as the `fake' sample.

In an unconditioned GAN model, there is no control over the data that is generated.  In the CGAN model, in contrast, by conditioning the model on additional information, it is possible to direct the data generation process \cite{CGAN}. Schematics of the GAN and CGAN models are depicted in Figure \ref{fig:gan}.

\makeatletter
\tikzset{
    dot diameter/.store in=\dot@diameter,
    dot diameter=2pt,
    dot spacing/.store in=\dot@spacing,
    dot spacing=9pt,
    dots/.style={
        line width=\dot@diameter,
        line cap=round,
        dash pattern=on 0pt off \dot@spacing
    }
}
\makeatother

\begin{figure*}
\centering
\tikzstyle{input_x}=[circle, minimum size = 1cm, thick, draw =blue!80, ,fill = blue!10,node distance = 16mm]

\tikzstyle{input_z}=[circle, minimum size = 1cm, thick, draw =cyan!80, ,fill = cyan!10,node distance = 16mm]
\tikzstyle{output}=[circle, minimum size = 1cm, thick, draw =red!80, ,fill = red!10,node distance = 16mm]
\tikzstyle{input}=[circle, minimum size = 1cm, thick, draw =green!80, ,fill = green!10,node distance = 16mm]

\tikzstyle{nod}=[circle, scale=0.5, thick, draw =gray!100, ,fill = gray!100]

\tikzstyle{disr}=[diamond, minimum size = 1.2cm, thick, draw =green!80, ,fill = green!10]
\tikzstyle{disf}=[diamond,minimum size = 1.2cm,  thick, draw =red!80, ,fill = red!10]

\tikzstyle{box}=[rectangle,fill = yellow!10, minimum width=3cm,rounded corners,thick,draw=yellow!100, inner sep=5mm, font=\sffamily]
\tikzstyle{box_d}=[rectangle,fill = black!100, text=white, minimum width=1cm,rounded corners,thick,draw=black!100, inner sep=2mm, font=\sffamily\bfseries]
\tikzstyle{box_r}=[rectangle,minimum width=1cm, inner sep=2mm, font=\sffamily]

\tikzstyle{dr} = [-stealth, thick, gray!100, line width=1pt]
\tikzstyle{-latex} = [-stealth, thick, gray!100, line width=1pt]

\resizebox{1\textwidth}{!}{
\begin{tikzpicture}

	\node[input_z] (z) {$\mathbf{z}_t$};
	\node[box, right=1cm of z] (gen) {Generator};
	
	\node[output, right=0.8cm of gen] (lgen) {$y_{t'}^{f}$};
	
	\node[input, above=1cm of lgen] (ly) {$y_{t'}$};
	\node[box, below right=0.05cm and 2cm of ly] (dis)	{\begin{tabular}{cc}
	GAN \\
	Discriminator \\
	\end{tabular} };
	\node[box_d, below right=0.05cm and 1cm of dis] (fake) {Fake};
	\node[box_d, above right=0.05cm and 1cm of dis] (real) {Real};
	
	\node[box_r, below=0.4cm of lgen] (lab) {a) GAN};

	\draw[dr] (gen) to (lgen);
	\draw[dr] (z) to (gen);

	\draw[-latex] (ly) -- ++(1,0) |- (dis);
	\draw[-latex] (lgen) -- ++(1,0) |- (dis);
	\draw[-latex] (dis) -- ++(2,0) |- (real);
	\draw[-latex] (dis) -- ++(2,0) |- (fake);
\end{tikzpicture}
}

\vspace{1.5cm}

\resizebox{1\textwidth}{!}{
\begin{tikzpicture}

	\node[input_x] (x) {$\mathbf{x}_t$};
	\node[input_z,  below=1cm of x] (z) {$\mathbf{z}_t$};
	\node[box, above right=0.0cm and 1cm of z] (gen) {Generator};
	
	\node[output, right=0.8cm of gen] (lgen) {$y_{t'}^{f}$};
	\node[nod, right=8cm of x] (n1) {};
	\node[nod, above=2.4cm of n1] (n2) {};
	
	\node[input_x, above=1.2cm of lgen] (x1) {$\mathbf{x}_t$};
	\node[input, above=1cm of x1] (ly) {$y_{t'}$};
	\node[box, below right=0.5cm and 1.5cm of n2] (dis) {\begin{tabular}{cc}
	CGAN \\
	Discriminator \\
	\end{tabular} };
	\node[box_d, below right=0.05cm and 1cm of dis] (fake) {Fake};
	\node[box_d, above right=0.05cm and 1cm of dis] (real) {Real};
	
	\node[box_r, below right=0.4cm and 0.4cm of lgen] (lab) {b) CGAN};

	\draw[dr] (x) to (n1);
	\draw[dr] (lgen) to (n1);	
	\draw[dr] (x1) to (n2);
	\draw[dr] (gen) to (lgen);
	\draw[dr] (ly) to (n2);

	\draw[-latex] (x) -- ++(1,0) |- (gen);
	\draw[-latex] (z) -- ++(1,0) |- (gen);
	\draw[-latex] (n1) -- ++(1,0) |- (dis);
	\draw[-latex] (n2) -- ++(1,0) |- (dis);	
	\draw[-latex] (dis) -- ++(2,0) |- (real);
	\draw[-latex] (dis) -- ++(2,0) |- (fake);

\end{tikzpicture}
}
\caption{\textit{Schematic of (a) GAN and (b) CGAN models, showing Generator and Discriminator components and associated variables.}}
\label{fig:gan}
\end{figure*}
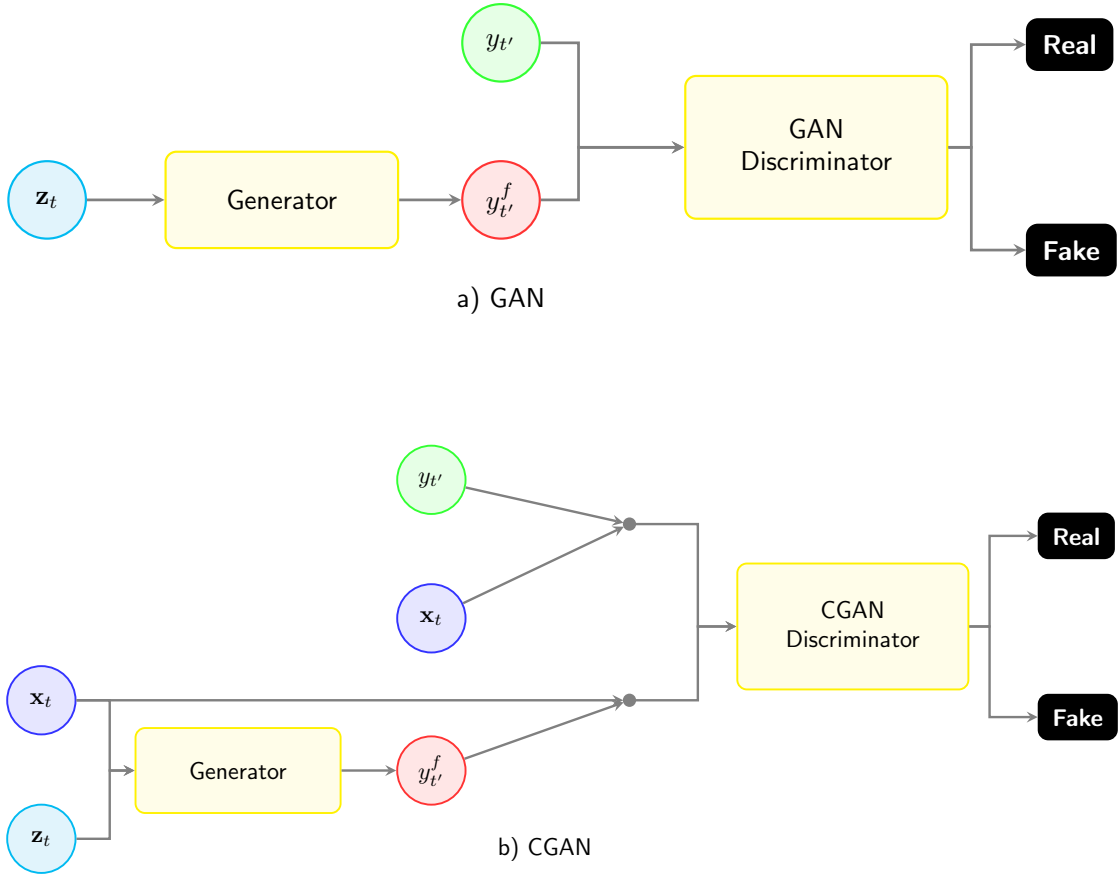

\section{The MD-CGAN Model Framework}\label{model}

As with the GAN and CGAN methods, we consider a time series, $y_t$. Our aim now is to infer the \emph{posterior distribution} over some $y_{t' > t}$, conditioned on the set of observations, $\mathbf{x}_t$.  In order to form the posterior distribution we model the \emph{full conditional density} $p(y_{t'} | \mathbf{x}_t)$ as an adversarial network.  To achieve this we use a Mixture Density Network (MDN) model \cite{bishop} for the generator $G$.  The inputs to the generator network are as per the CGAN approach, $\mathbf{x}_t$ and $\mathbf{z}_g$, where $\mathbf{z}_g$ is, as before, a collection of samples from a normal distribution, $p(z_n)_g = \mathcal{N}(0,\mathrm{var}_{\mathrm{data}})$.  The outputs of  $G_{t'}(\mathbf{x}_t,\mathbf{z}_g)$ are now, however, the parameters of the Gaussian mixture posterior over the forecast. This mixture has mixing coefficient, standard deviation and mean for the $i$-th component denoted as $\alpha_{i}$, $\sigma_{i}$, and $\mu_{i}$ respectively.  As first proposed in \cite{bishop}, we achieve this by using latent variables $\mathbf{s} = \{ \mathbf{s}_{\alpha}, \mathbf{s}_{\sigma}, \mathbf{s}_{\mu} \}$, conditioned on the inputs. The mapping from $[\mathbf{x}_t,\mathbf{z}_g] \mapsto  \mathbf{s} \mapsto \{ \alpha_i, \sigma_i, \mu_{i} \}$ is modelled via our network.  As the mixings must satisfy $\sum_{i}\alpha_i=1$, we map $\mathbf{s}_{\alpha}$ to $\mathbf{\alpha}$ via the \emph{softmax} function, where $\mathbf{\alpha}_i=\frac{\exp({s}_{\alpha,i})}{\sum \limits_{i'} \exp({s}_{\alpha, i'})}$.  The elements of $\mathbf{\sigma}$ are strictly positive so we adopt, $\sigma_i = \exp(s_{\sigma,i})$. Finally the means can be mapped directly from the latent variables, hence $\mu_i= s_{\mu,i}$. Schematically, the MD-CGAN method is depicted in Figure \ref{chart}.

\makeatletter
\tikzset{
    dot diameter/.store in=\dot@diameter,
    dot diameter=2pt,
    dot spacing/.store in=\dot@spacing,
    dot spacing=9pt,
    dots/.style={
        line width=\dot@diameter,
        line cap=round,
        dash pattern=on 0pt off \dot@spacing
    }
}
\makeatother

\begin{figure*}
\centering
\tikzstyle{input_x}=[circle, minimum size = 1cm, thick, draw =blue!80, ,fill = blue!10,node distance = 16mm]

\tikzstyle{input_z}=[circle, minimum size = 1cm, thick, draw =cyan!80, ,fill = cyan!10,node distance = 16mm]
\tikzstyle{output}=[circle, minimum size = 1cm, thick, draw =red!80, ,fill = red!10,node distance = 16mm]
\tikzstyle{input}=[circle, minimum size = 1cm, thick, draw =green!80, ,fill = green!10,node distance = 16mm]

\tikzstyle{nod}=[circle, scale=0.5, thick, draw =gray!100, ,fill = gray!100]

\tikzstyle{disr}=[diamond, minimum size = 1.2cm, thick, draw =green!80, ,fill = green!10]
\tikzstyle{disf}=[diamond,minimum size = 1.2cm,  thick, draw =red!80, ,fill = red!10]

\tikzstyle{box}=[rectangle,fill = yellow!10, minimum width=3cm,rounded corners,thick,draw=yellow!100, inner sep=5mm, font=\sffamily]
\tikzstyle{box_d}=[rectangle,fill = black!100, text=white, minimum width=1cm,rounded corners,thick,draw=black!100, inner sep=2mm, font=\sffamily\bfseries]
\tikzstyle{box_r}=[rectangle,minimum width=1cm, inner sep=2mm, font=\sffamily]

\tikzstyle{dr} = [-stealth, thick, gray!100, line width=1pt]
\tikzstyle{-latex} = [-stealth, thick, gray!100, line width=1pt]

\resizebox{1\textwidth}{!}{
\begin{tikzpicture}

	\node[input_x] (x) {$\mathbf{x}_t$};
	\node[input_z,  below=1cm of x] (z) {$\mathbf{z}_t$};
	\node[box, above right=0.0cm and 1cm of z] (gen) {Generator};
	\node[draw, draw =red!80, ,fill = red!10, rectangle, align=left ,rounded corners,thick, right=0.8 of gen](pr){ $\mathbf{\alpha}_i$ \\ $\mathbf{\sigma}_i$ \\ $\mathbf{\mu}_i$ };
	
	\node[disf, right=0.8cm of pr] (lgen) {$\mathcal{L}(G_{t'})$};
	\node[nod, right=10cm of x] (n1) {};
	\node[nod, above=2.4cm of n1] (n2) {};
	
	\node[input_x, above=1.2cm of pr] (x1) {$\mathbf{x}_t$};
	\node[input, above=1cm of x1] (y) {$y_{t'}$};
	\node[disr, right=0.5cm of y] (ly) {$\mathcal{L}(y_{t'})$};
	
	\node[box, below right=0.5cm and 1.5cm of n2] (dis) {\begin{tabular}{cc}
	MD-CGAN \\
	Discriminator \\
	\end{tabular} };

	\node[box_d, below right=0.05cm and 1cm of dis] (fake) {Fake};
	\node[box_d, above right=0.05cm and 1cm of dis] (real) {Real};

	
	\draw[dr] (gen) to (pr);
	\draw[dr] (pr) to (lgen);
	\draw[dr] (x) to (n1);
	\draw[dr] (lgen) to (n1);	
	\draw[dr] (x1) to (n2);
	\draw[dr] (ly) to (n2);
	\draw[dr] (y) to (ly);

	\draw[-latex] (x) -- ++(1,0) |- (gen);
	\draw[-latex] (z) -- ++(1,0) |- (gen);
	\draw[-latex] (n1) -- ++(1,0) |- (dis);
	\draw[-latex] (n2) -- ++(1,0) |- (dis);	
	\draw[-latex] (dis) -- ++(2,0) |- (real);
	\draw[-latex] (dis) -- ++(2,0) |- (fake);

\end{tikzpicture}
}
\caption{\textit{Schematic of the MD-CGAN model.  We note that the discriminator in the MD-CGAN model has a different loss function and structure to the GAN and CGAN models.}}
\label{chart}
\end{figure*}
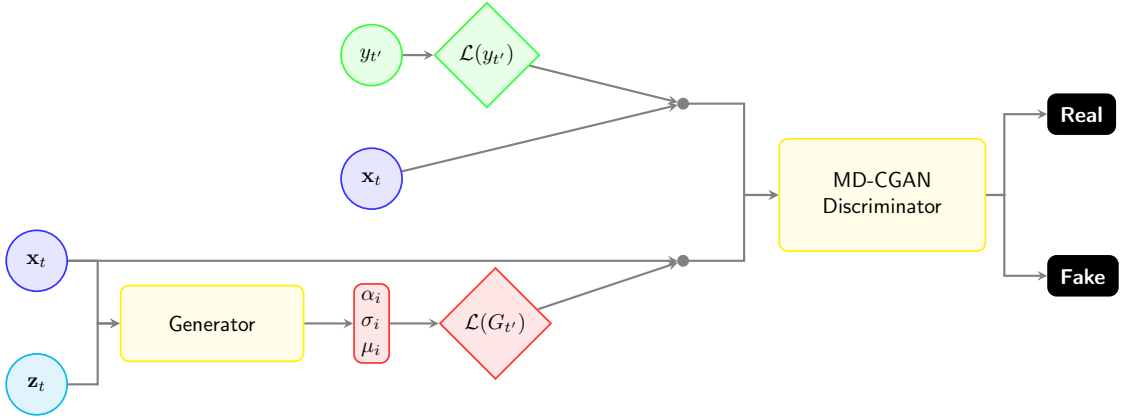

The above formalism allows us to directly model the predictive likelihood conditioned on an input, and the likelihood of $G$, conditioned on the observations $\mathbf{x}_t$ and samples $\mathbf{z}_g$ as:
\begin{equation}
\mathcal{L}(G_{t'}(\mathbf{x}_t,\mathbf{z}_g))= \sum_{i=1}^{m} \alpha_i(\mathbf{x}_t,\mathbf{z}_g) 
\mathcal{N}_i(y_{t'}|\mu_{i}(\mathbf{x}_t,\mathbf{z}_g), \sigma_{i}(\mathbf{x}_t,\mathbf{z}_g))
\label{eq1}
\end{equation} 
where $m$ is the number of mixture components.

As in the CGAN model, the discriminator, $D$, is also conditioned on $\mathbf{x}_t$. The input to the discriminator model is, by design, $\mathbf{x}_t \sqrt{2 \pi} \sigma_{a} \mathcal{L}(y_{t'})$, where $\sigma_a$ is the standard deviation of the set of observed $y_t$ \footnote{We note that the GAN approach is not sensitive, within reason, to this value (it is, in effect, a constant in the update equations) and we discuss its setting later in the paper.}.  For true values of $y_{t'}$, $\sqrt{2 \pi} \sigma_{a} \mathcal{L}(y_{t'})$ is maximized.  The generator tries to `fool' the discriminator by generating $G_{t'}$ such that the $\sqrt{2 \pi} \sigma_{a} \mathcal{L}(G_{t'})$ is maximized.  The loss function for the generator, $L_G$ is as in Equation \ref{gnerator_loss},  The discriminator network, on the other hand, attempts to differentiate between true $y_{t'}$ values and the pseudo-values created by the generator.  The loss function for the discriminator, $L_D$ is as in Equation \ref{disc_loss}. We note that the lowest value of the discriminator loss is achieved when $\sqrt{2 \pi} \sigma_{a} \mathcal{L}(y_{t'})$ is maximal (unity) and $\mathcal{L}(G_{t'}(\mathbf{x}_t,\mathbf{z}_g))$ is minimal (zero).
\begin{equation}
L_G = \mathbb{E}_{z\sim P_z (z)}[-\mathcal{L}(G_{t'}(\mathbf{x}_t,\mathbf{z}_g))]
\label{gnerator_loss}
\end{equation}
\begin{align}
L_D = & \mathbb{E}_{y\sim P_{\mathrm{data}}(y)}[\Vert \mathbf{x}_t \sqrt{2 \pi} \sigma_{a} \mathcal{L}
(y_{t'}) - \mathbf{x}_t \Vert ^2]+ \nonumber \\
&\mathbb{E}_{z\sim P_z (z)}[\Vert \mathbf{x}_t \sqrt{2 \pi} \sigma_{a} \mathcal{L}(G_{t'}(\mathbf{x}_t,\mathbf{z}_g))\Vert ^2]
\label{disc_loss}
\end{align}
Our algorithm thus follows the steps in Algorithm \ref{algo} below.
\begin{algorithm}
  \caption{MD-CGAN Algorithm}\label{algo}
  \begin{algorithmic}[1]
      \For{number of training iterations}
            \For{$j$ steps do}

		        \State Sample $N$ noise samples, \{$\mathbf{z}^1$,...,$\mathbf{z}^N$\} from $p_g(\mathbf{z})$
		        \State Sample $N$ data points, \{$\mathbf{x}^1$,...,$\mathbf{x}^N$\} from $p_\mathrm{data}(\mathbf{x})$
		        \State Update the discriminator by descending its stochastic gradient: \par
				\vspace*{-2em} 
\begin{align*}
	\nabla_{\theta_{\mathcal{L}}} \sum_{n=1}^{N} &[\Vert \mathbf{x}^{(n)} \sqrt{2 \pi} \sigma_{a} \mathcal{L}(y^{(n)} ) - \mathbf{x}^{(n)} \Vert ^2+ \\
	& \Vert \mathbf{x}^{(n)} \sqrt{2 \pi} \sigma_{a} \mathcal{L}(G(\mathbf{z}^{(n)}, \mathbf{x}^{(n)})) \Vert ^2]
\end{align*}				
	      \EndFor
			\State Sample $N$ noise samples, \{$\mathbf{z}^1$,...,$\mathbf{z}^N$\} from $p_g(\mathbf{z})$
			\State Update the generator by descending its stochastic gradient: \par
				\vspace*{-2em} 
				\begin{align*}
				\nabla_{\theta_g} \sum_{n=1}^{N} -  \mathcal{L}(G(\mathbf{z}^{(n)}, \mathbf{x}^{(n)}) )
				\end{align*}				
      \EndFor
  \end{algorithmic}
\end{algorithm}

\section{Experiments}\label{experiments}

\subsection{Comparison with other Learning Models}\label{sub1}

To provide a range of comparisons to methods related  to this work, we compare our MD-CGAN model to the following baseline methods: the Mixture Density Network model (MDN), chosen to baseline mixture density outputs against \cite{bishop}; the CGAN model, chosen as a well-known GAN approach \cite{CGAN}; and a ``standard'' Multi-Layer Perceptron (MLP) neural network (SNN) as a simple,  yet  effective,  baseline.  As a more traditional, parametric, benchmarking model we use regular (linear-Gaussian) Auto-Regressive (AR) models \cite{Papoulis}, with parameters obtained by standard least-squares maximum-likelihood estimation.
To promote as fair comparison as possible for the nonlinear methods, we use a \emph{common} neural network architecture across core components in the models, choosing the neural network structure commonly used for GAN models in recent literature \cite{CGAN}.  We do not alter this structure throughout our experiments and we keep to fixed hyperparameter settings (guided by those used in \cite{CGAN}).  Figure \ref{fig:nnstruct} provides a schematic of the structure used.  We note that, whilst the lengths of the input and output vectors are dependent on the model, the structure of all networks remains constant.

\subsection{Details of implementation}

All models were coded in the Python language and we use the Keras library \cite{Keras} to build the neural networks.

\begin{figure*}[!h]
\begin{adjustwidth}{-4em}{} 

\tikzstyle{box}=[rectangle,fill = yellow!10, minimum width=4.5cm,rounded corners,thick,draw=yellow!100, inner sep=2mm, font=\sffamily\Large]
\tikzstyle{box_d}=[rectangle,fill = cyan!10, minimum width=7.5cm,rounded corners,thick,draw=cyan!100, inner sep=2mm, font=\sffamily\Large]
\tikzstyle{box_op}=[rectangle,fill = brown!20, minimum width=7cm, inner sep=2mm, font=\sffamily\Large]
\tikzstyle{box_s_d}=[rectangle,fill = blue!10, minimum width=6cm, inner sep=2mm, font=\sffamily\Large]
\tikzstyle{box_s_LR}=[rectangle,fill = red!10, minimum width=6cm, inner sep=2mm, font=\sffamily\Large]
\tikzstyle{box_s_DO}=[rectangle,fill = magenta!10, minimum width=6cm, inner sep=2mm, font=\sffamily\Large]
\tikzstyle{box_s_BN}=[rectangle,fill = orange!10, minimum width=6cm, inner sep=2mm, font=\sffamily\Large]

\tikzstyle{box_d_o}=[rectangle,fill = black!100, text=white, minimum width=1cm,rounded corners,thick,draw=black!100, inner sep=2mm, font=\sffamily\bfseries\Large]

\tikzstyle{box_r}=[rectangle,fill = gray!10,minimum width=1cm, rounded corners,thick,draw=gray!10,inner sep=2mm, font=\sffamily\Large]

\tikzstyle{dr} = [-stealth, thick, gray!100, line width=1pt]
\tikzstyle{-latex} = [-stealth, thick, gray!100, line width=1pt]

\resizebox{1.2\textwidth}{!}{
\begin{tikzpicture}

	\node[box_r] at (-8, 2.5) {
Structure of SNN, MDN and GAN model generators};

	\node[box_r] at (11, 2.5) {
Structure of GAN model Discriminators};

	\node[box] (MD-CGAN) {
\begin{tabular}{cc}
	MD-CGAN Input: \\
		$\mathbf{x}$, $\mathbf{z}$ \\
	\end{tabular}	
};

	\node[box, left=0.7cm of MD-CGAN] (MDN) {
	\begin{tabular}{cc}
	MDN Input: \\
		$\mathbf{x}$ \\
	\end{tabular}	
};

	\node[box, left=0.7cm of MDN] (CGAN) {
\begin{tabular}{cc}
	CGAN Input: \\
		$\mathbf{x}$, $\mathbf{z}$ \\
	\end{tabular}	
};

	\node[box, left=0.7cm of CGAN] (SNN) {
	\begin{tabular}{cc}
	SNN Input: \\
		$\mathbf{x}$ \\
	\end{tabular}	
};

	\node[box_s_d] at (-8, -4)(DL) {Dense layer $n$};
	\node[box_s_LR, below=0cm of DL](LR) {Leaky ReLU};
	\node[box_s_DO, below=0cm of LR](DO) {Dropout};
	\node[box_s_BN, below=0cm of DO](BN) {Batch Normalization};

	\node[box_s_d] at (-8, -8.5)(DL1) {Dense layer $2n$};
	\node[box_s_LR, below=0cm of DL1](LR1) {Leaky ReLU};
	\node[box_s_DO, below=0cm of LR1](DO1) {Dropout};
	\node[box_s_BN, below=0cm of DO1](BN1) {Batch Normalization};

	\node[box_s_d] at (-8, -13)(DL2) {Dense layer $4n$};
	\node[box_s_LR, below=0cm of DL2](LR2) {Leaky ReLU};
	\node[box_s_DO, below=0cm of LR2](DO2) {Dropout};
	\node[box_s_BN, below=0cm of DO2](BN2) {Batch Normalization};

	\node[box, below=17cm of MD-CGAN] (MD-CGAN_o) {
\begin{tabular}{cc}
	MD-CGAN Output: \\
		${\{\alpha_i, \sigma_i, \mu_i \}}$ \\
	\end{tabular}	
};

	\node[box, left=0.7cm of MD-CGAN_o] (MDN_o) {
\begin{tabular}{cc}
	MDN Output: \\
		${\{\alpha_i, \sigma_i, \mu_i \}}$ \\
	\end{tabular}	
};

	\node[box, left=0.7cm of MDN_o] (CGAN_o) {
\begin{tabular}{cc}
	CGAN Output: \\
		$y^f_{t'}$ \\
	\end{tabular}	
};

	\node[box, left=0.7cm of CGAN_o] (SNN_o) {
\begin{tabular}{cc}
	SNN Output: \\
		$y^f_{t'}$ \\
	\end{tabular}	
};

	\draw[-latex] (SNN) -- ++(0,-1.5) -| (DL);
	\draw[-latex] (CGAN) -- ++(0,-1.5)-| (DL);
	\draw[-latex] (MDN) -- ++(0,-1.5)-| (DL);
	\draw[-latex] (MD-CGAN) -- ++(0,-1.5)-| (DL);
	
	\draw[dr] (BN) to (DL1);
	\draw[dr] (BN1) to (DL2);

	\draw[-latex] (BN2) -- ++(0,-2) -| (SNN_o);
	\draw[-latex] (BN2) -- ++(0,-2) -| (CGAN_o);
	\draw[-latex] (BN2) -- ++(0,-2) -| (MDN_o);
	\draw[-latex] (BN2) -- ++(0,-2) -| (MD-CGAN_o);

	\node[box_d, right=1.5cm of MD-CGAN] (CGAN des) {
	\begin{tabular}{cc}
	CGAN Input: \\
	$\{\mathbf{x},y_{t'}\}$ , $\{\mathbf{x},y^{f}_{t'}\}$\\
	\end{tabular}	
	};

	\node[box_d, below=0.7cm of CGAN des] (CGAN des_1) {
	$[\mathbf{x},y_{t'}]$ , $[\mathbf{x},y^{f}_{t'}]$	};

	\node[box_d, right=0.7cm of CGAN des] (MD-CGAN des) {
	\begin{tabular}{cc}
	MD-CGAN Input: \\
			$\{\mathbf{x},\mathcal{L}(y_{t'})\}$ , $\{\mathbf{x},\mathcal{L}(G_{t'})\}$ \\
	\end{tabular}	
};

	\node[box_d, below=0.7cm of MD-CGAN des] (MD-CGAN des_1) {
$\mathbf{x}_t \sqrt{2 \pi} \sigma_{a} \mathcal{L}(y_{t'})$ , 
$\mathbf{x}_t \sqrt{2 \pi} \sigma_{a} \mathcal{L}(G_{t'})$ };

	\node[box_s_d] at (11.5, -4.5)(DL_d) {Dense layer $2n$};
	\node[box_s_LR, below=0cm of DL_d](LR_d) {Leaky ReLU};
	\node[box_s_DO, below=0cm of LR_d](DO_d) {Dropout};

	\node[box_s_d] at (11.5, -8.25)(DL_d_1) {Dense layer $2n$};
	\node[box_s_LR, below=0cm of DL_d_1](LR_d_1) {Leaky ReLU};
	\node[box_s_DO, below=0cm of LR_d_1](DO_d_1) {Dropout};

	\node[box_s_d] at (11.5, -12)(DL_d_2) {Dense layer $2n$};
	\node[box_s_LR, below=0cm of DL_d_2](LR_d_2) {Leaky ReLU};
	\node[box_s_DO, below=0cm of LR_d_2](DO_d_2) {Dropout};

	\node[box_op, below=14.5cm of CGAN des] (CGAN des_o) {(C)GAN loss function};
	\node[box_op, below=14.5cm of MD-CGAN des] (MD-CGAN des_o) {MD-GAN loss function};

	\node[box_d_o]at (6, -19) (fake) {Fake};
	\node[box_d_o]at (9, -19) (real) {Real};

	\node[box_d_o]at (14.25, -19) (fake_1) {Fake};
	\node[box_d_o]at (17.25, -19) (real_1) {Real};

	\draw[dr] (CGAN des) to (CGAN des_1);
	\draw[dr] (MD-CGAN des) to (MD-CGAN des_1);

	\draw[-latex] (CGAN des_1) -- ++(0,-1) -| (DL_d);
	\draw[-latex] (MD-CGAN des_1) -- ++(0,-1)-| (DL_d);

	\draw[dr] (DO_d) to (DL_d_1);
	\draw[dr] (DO_d_1) to (DL_d_2);

	\draw[-latex] (DO_d_2) -- ++(0,-1.25) -| (CGAN des_o);
	\draw[-latex] (DO_d_2) -- ++(0,-1.25) -| (MD-CGAN des_o);
	
	\draw[-latex] (CGAN des_o) -- ++(0,-2) -| (fake);
	\draw[-latex] (CGAN des_o) -- ++(0,-2) -| (real);

	\draw[-latex] (MD-CGAN des_o) -- ++(0,-2) -| (fake_1);
	\draw[-latex] (MD-CGAN des_o) -- ++(0,-2) -| (real_1);

\end{tikzpicture}
}
\caption{\textit{Common Neural Network structure used across all models.}}
\label{fig:nnstruct}
\end{adjustwidth}
\end{figure*}
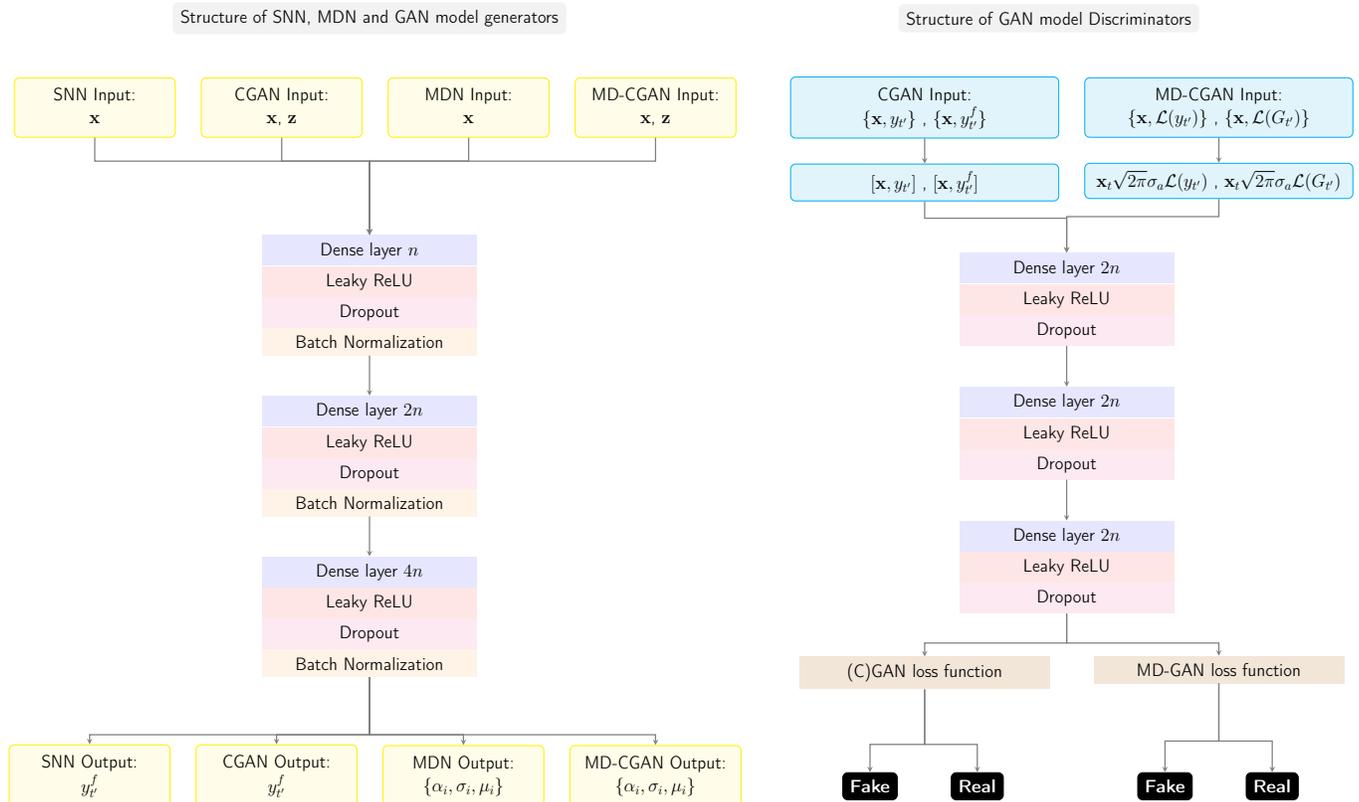

The neural networks in Figure \ref{fig:nnstruct} follow the structure of the CGAN model detailed in \cite{CGAN}. The hyperparameters in the models were set as follows: for the discriminator, both in CGAN and MD-CGAN, the dropout parameter is set to 0.4, and alpha for the leaky ReLU is set to 0.2.  For the generator the dropout rate is set to 0.5 and alpha for the leaky ReLU is, again, 0.2.  The parameter governing the number of neurons, $n$,  in the dense layers of the neural network modules is set to 20.  The variance parameter, $\sigma_a$, in the MD-CGAN models is set to 0.2.
During training, we follow the steps of algorithm 1, and set $j$ to 1 for all datasets. 

The number of training iterations is, however, specific to model and dataset.  With the exception of the GAN models we monitor the errors, until they reach saturation during training.  For the  GAN models, in which both the generator and the discriminator have a loss value per iteration, we monitor the average sample error from the training data and continue iteration until the errors reach saturation.

In all models we optimize parameters using the Adam optimizer \cite{adam}, with learning rate set to 0.001, exponential decay for the first momentum set to 0.9, exponential decay for the second momentum set to 0.999, and epsilon set to $10^{-7}$.

\subsection{Data}

We perform experiments on ten datasets with differing provenance; including Mackey-Glass chaotic dataset \cite{mg}; the Sunspot dataset \cite{sunspot} and eight financial time series. We first look at the one-step forecasting and the issue of (test set) noise resilience, focusing on controlled experiments with the Mackey-Glass and  Sunspot datasets (Subsection \ref{sec:data}).  We then expand our experiments to consider the financial datasets and increased forecast horizon in Subsection \ref{sec:finance}.

For all datasets we use, the time series is split into training and out-of-sample test sets. The training data sets in all our experiments comprise 2000 samples and all test sets consist of 400 sequential data points post the training set. All performance metrics are obtained from the test set only.  All nonlinear algorithms are provided, as input, the last $k$ data points, set to $k=5$ for the purpose of all our experiments. For the linear AR models, we run both an AR(5) model (corresponding to $k=5$) and, as a martingale baseline, the AR(0) model, in which the forecast is simple the previous observation. We note, however, that the value of $k=5$ is not optimized, and is chosen merely to allow simple comparisons across methods.  All data sets are pre-normalized to the [0,1] interval, again to allow for simpler comparison across data and methods.  We further note that both the CGAN and SNN models make \emph{point estimate} predictions, whilst MD-CGAN and MDN estimate posterior \emph{distributions}. To enable a simple comparison across models we therefore report the mean-square error (MSE) for all methods.  The number of mixture components, $m$, in both MD-CGAN and MDN is set to unity (we vary this in Section \ref{sec:vary_m}), to further ease comparison. The most-likely value (which for $m=1$ is merely the posterior mean) of the predictive distribution is taken as the forecast value for both the MDN and MD-CGAN models for the purposes of point-prediction error reporting.

\subsection{One-step forecasting}
\label{sec:data}
\begin{figure*}[!h]
\centering
\includegraphics[width=.45\textwidth, height=0.3\textwidth]{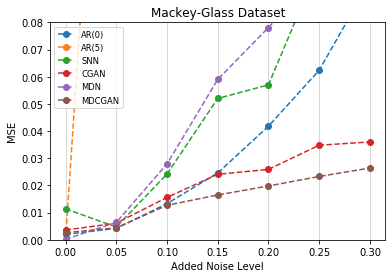}\hspace{0.5cm}
\includegraphics[width=.45\textwidth, height=0.3\textwidth]{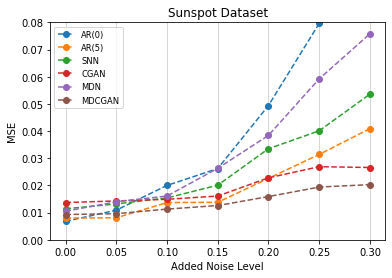}
\caption{Comparative MSE plots with increasing test set noise perturbation. We note that the GAN-based methods, particularly MD-CGAN, perform consistently even as noise levels increase.}
\label{errorbar}
\end{figure*}

\noindent \emph{Mackey-Glass and Sunspot time series:}
We start our experiments looking at one-step ahead forecasts on two well known data sets, the Mackey-Glass chaotic time series \cite{mg} and the Sunspot data set \cite{sunspot}. We consider one-step forecast errors in the presence of increasing test set noise. We add 5\% to 30\% (by amplitude) normally distributed noise to the test data (from a GAN perspective, these input perturbations are, in effect, treated as adversarial attacks).  We note that no noise is added to the training dataset.  Mean Square Errors (MSE) are presented in Tables \ref{table1} and \ref{table2} and Figure \ref{errorbar} for all algorithms considered.  For both datasets we see that MD-CGAN has the best performance for noise levels of 10\% and above.  Indeed we see that the GAN models (particularly MD-CGAN) perform consistently well (especially in the Mackey-Glass example) across multiple noise levels, indicating that the approach is particularly resilient to additive observation noise. This is to be expected, as GAN approaches treat the additive noise as adversarial perturbation to the input, against which they are designed to be robust.

In the next section we investigate model performance at longer forecast horizons in the finance domain, which is known to contain variable amounts of stochasticity. Financial times series are often dominated by stochastics, and we expect GAN approaches to be well-suited to forecasting in these circumstances.

\begin{table*}[t]
\small
\caption{Mackey-Glass data: MSE variation with added noise level.}
\vspace{0cm}
\begin{adjustwidth}{-0.8cm}{}
        \begin{tabular}{l c c c c c c c}
            \toprule
             
                &  \textbf{0\% noise} & \textbf{5\% noise} & \textbf{10\% noise} & \textbf{15\% noise}&   \textbf{20\% noise}& \textbf{25\% noise} & \textbf{30\% noise}\\
               
                \cmidrule{2-8}

	AR(0) & 0.0020 & \textbf{0.0042}  & 0.0133 &  0.0246 	& 0.0418    & 0.0624  & 0.0951 \\
	
	AR(5) & \textbf{4.1e-06} & 0.2794&  1.1558 &   2.6581  &  4.4868  &  7.3152 & 10.1527 \\

	SNN & 0.0014  & 0.0047  &  0.0242 &   0.0519  & 0.0570   &  0.1013  & 0.1640 \\

	CGAN & 0.0036 &  0.0061 &  0.0155 &  0.0240  & 0.0259   &  0.0347  &  0.0360\\

	MDN & 0.0002 &  0.0064&  0.0278 & 0.0589  &0.0780   &0.0980  &  0.1402\\

	MD-CGAN & 0.0026  &  0.0044 & \textbf{0.0126} & \textbf{0.0165}  &\textbf{0.0197} &\textbf{0.0233} &\textbf{ 0.0264 }\\
    
            \bottomrule
        \end{tabular}
        \end{adjustwidth}
        
        \label{table1}
\end{table*}

\begin{table*}[t]
\small
\caption{Sunspot data: MSE variation with added noise level.}
\vspace{0cm}
\begin{adjustwidth}{-0.8cm}{}
        \begin{tabular}{l c c c c c c c}
            \toprule
             
                &  \textbf{0\% noise} & \textbf{5\% noise} & \textbf{10\% noise} & \textbf{15\% noise}&   \textbf{20\% noise}& \textbf{25\% noise} & \textbf{30\% noise}\\
               
                \cmidrule{2-8}

	AR(0) & 0.0080  & 0.0109 & 0.0200  &  0.0262  &  0.0494  & 0.0795  & 0.0974 \\
	
	AR(5) & \textbf{ 0.0068} &  \textbf{0.0081 }  &  0.0137 &  0.0138 &  0.0226 & 0.0314 & 0.0409 \\
	
	SNN & 0.0114 & 0.0132 & 0.0154 &   0.0201 & 0.0335  & 0.0401 & 0.0536 \\
	
	CGAN & 0.0137  &  0.0143 & 0.0149 &   0.0161 &  0.0228 & 0.0269 & 0.0266 \\
	
	MDN & 0.0105 & 0.0140 & 0.0161 & 0.0263 & 0.0384 & 0.0592 & 0.0758 \\
	
	MD-CGAN & 0.0093 & 0.0096 & \textbf{0.0113} & \textbf{ 0.0126} & \textbf{0.0159} & \textbf{0.0194} &
\textbf{0.0203} \\
	    
            \bottomrule
        \end{tabular}
        \end{adjustwidth}
        
        \label{table2}
\end{table*}

\subsection{Financial forecasts over longer-horizons}
\label{sec:finance}
One-step forecasts were presented in Subsection \ref{sub1}.  Here we extend analysis of the financial data over longer horizons.
All models were used to make estimates over a horizon of ten weeks for an extended set of financial time series, namely: US initial jobless claims (USIJC, weekly intervals, \cite{jobless});, EURUSD foreign exchange daily rates (EURUSD FX rate, \cite{eurusd}), WTI crude oil spot prices (WTI, \cite{energy}), Henry Hub Natural Gas spot prices (Nat Gas, \cite{energy}), CBOE Volatility Index (VIX \cite{vix}), New York Harbor No. 2 Heating Oil Spot Price (Heating Oil \cite{energy}), Invesco DB US Dollar Index Bullish Fund (USD Index \cite{UUP}), and iShares MSCI Brazil Small-Cap ETF (EM ETF \cite{EWZS}).  We forecast over a ten-week horizon, representing a 50 step forecast for the daily datasets (FX, WTI, Nat Gas, VIX, Heating Oil, USD Index \& EM ETF), and 10 steps for the weekly USIJC dataset. Our comparisons, as previously, include standard econometric linear models, namely the 5-th order autoregressive,  AR(5), model and the martingale, or AR(0) model, in which the forecast is the last observed datum. Taking the martingale model as a baseline, we present in Table \ref{tableRatio} the mean-square errors as the ratio to the martingale model error. We note that the MD-CGAN approach delivers ratios below unity and provides the lowest error of almost all models in this scenario.
\begin{table*}[t]
\small
\caption{Ratio of model MSE to martingale, AR(0), baseline model over 10-week forecast horizon.}
    \begin{adjustwidth}{-2cm}{}
        \begin{tabular}{l c c c c c   c c c}
            \toprule
             
                &     \textbf{USIJC}& \textbf{EURUSD FX rate} & \textbf{WTI}& \textbf{Nat Gas} & \textbf{VIX index} &\textbf{Heating Oil} &\textbf{USD Index} &\textbf{EM ETF}\\
               
                \cmidrule{2-9}
    AR(5)  &0.78 & 1.91 &0.85 &1.01 & 0.71 &0.82 & 1.24&0.89\\           
	SNN &0.79 & 1.25 &0.89&0.94&0.71 & 0.93&1.34 &0.82\\
	CGAN &0.77 & 0.85 &1.53&1.07&0.91 &\textbf{0.54} &1.37 & 0.69\\
	MDN &0.84 & 3.48 &1.48 &1.13 & 0.77 & 0.89&0.68 &0.81\\
	MD-CGAN &\textbf{0.73} & \textbf{0.76}  &\textbf{0.80} &\textbf{0.82} & \textbf{0.66} & 0.59& \textbf{0.54}&\textbf{0.65}\\
    
            \bottomrule
        \end{tabular}
        \end{adjustwidth}
        
        \label{tableRatio}
\end{table*}

\subsection{Multi-modal posterior predictions}
\label{sec:vary_m}
\begin{figure*}[!h]
\centering
\includegraphics[width=.43\textwidth, height=0.3\textwidth]{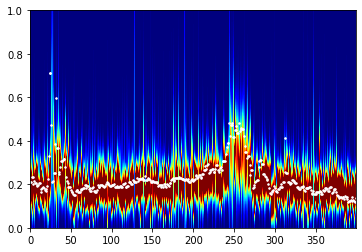}\hspace{0.5cm}
\includegraphics[width=.43\textwidth, height=0.3\textwidth]{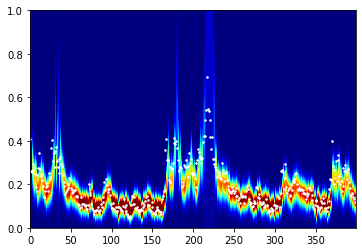}
\caption{Estimated distributions for (left) Nat Gas with $m=2$ and (right) VIX index with $m=1$ over out-of-sample test datasets.  True samples are shown as white dots.  Red indicates high data likelihood and blue data low likelihood under the MD-CGAN model.}
\label{figDisPlot}
\end{figure*}

Finally, we compare the performance of MD-CGAN over varying numbers of mixture components.  In all the previous experiments we set $m=1$ (hence the model produced a single predictive Gaussian posterior).  Here we briefly present the results for the five finance datasets with $m \in \{1, 2, 3\}$.  We report (negative) log-likelihood measures (as we do not compare against point-value models in this section) and consider one-step forecasts on all the data sets. Table \ref{table3} presents the performance across data sets for varying numbers of mixture components in the posterior prediction.  Figure \ref{figDisPlot} shows the predicted distribution for the test datasets for Nat Gas and Vix Index.

We note performance improvement for some datasets for $m>1$. We do not attempt to infer $m$, though choosing its value based on performance on a set of cross-validation data would be an option, as would enforcing regularization over the mixture model posterior, through more extensive use of Bayesian inference. We leave these extensions for future research.

\begin{table*}[t]
\small
\caption{Negative log-likelihood variation with $m$, with standard deviations in brackets.}
\vspace{0cm}
    \begin{adjustwidth}{-1.5cm}{}
        \begin{tabular}{l c c c  c c c c c}
            \toprule
             
                &       \textbf{USIJC}& \textbf{EURUSD FX rate} & \textbf{WTI} & \textbf{Nat Gas}& \textbf{VIX index}& \textbf{Heating Oil}& \textbf{USD index}& \textbf{EM ETF}\\
               
                \cmidrule{2-9}
	$m=1$ &  -1.01 & \textbf{-1.79}&\textbf{-1.75}& -1.28 & \textbf{-1.50} &\textbf{-1.63} & -0.65&\textbf{-1.26}\\
	 &(0.65)&(0.42)&(0.27)& (0.72)  & (0.94) & (0.23) &(0.38)&(0.41) \\
	$m=2$ &  -1.05 & -0.98  &-1.24&\textbf{-1.33} &-1.39 &-1.37 & -0.83&-1.10\\
	&(0.39)&(0.33)&(0.32)& (0.66)& (0.78)&(0.24) &(0.32)&(0.32) \\
    $m=3$ &  \textbf{-1.09} & -0.67&-1.33& -1.25 & -1.48 & -1.35&\textbf{-0.86} &-1.09\\
	&(0.34)&(0.18)&(0.37)& (0.63)& (0.92)& (0.26) &(0.12)&(0.27)\\
            \bottomrule
        \end{tabular}
        \end{adjustwidth}
        
        \label{table3}
\end{table*}

\section{Conclusion} \label{conclusion}
In this paper we present the MD-CGAN model, which offers extensions to the CGAN \cite{CGAN} methodology, particularly to allow for GAN inference of a (multi-modal) posterior distribution over forecast values. In the experiments considered, we find the MD-CGAN approach outperforms other methods on all datasets in which noise is prevalent, including all the financial time series investigated, over long term forecast horizons.
As a GAN model, our approach retains adversarial robustness in forecasting, which we find is most notable when noise is extensively present in data. Our method is thus particularly well suited to dealing with financial data. Furthermore,  MD-CGAN can effectively estimate a flexible posterior distribution, in contrast to standard GAN models which (almost without exception) produce point value outputs.  Exploiting this rich, multi-model, posterior distribution is not reported in detail here but will feature in follow-up work. In summary, the MD-CGAN model combines the advantageous features of both flexible probabilistic forecasting and GAN methods. We see this as a particularly useful approach for dealing with time series in which noise is significant and for providing robust, long-term forecasts beyond simple point estimates.




\bibliography{mybibfile}

\end{document}